\documentclass[final,twocolumn,10pt,conference]{IEEEtran}
\usepackage{graphicx,times,amsmath,amssymb,cite,subfigure}
\IEEEoverridecommandlockouts \allowdisplaybreaks
\DeclareMathOperator*{\argmin}{\arg\!\min}
\usepackage[lined,ruled,commentsnumbered]{algorithm2e}

\begin{document}

\title{Active Semi-supervised Transfer Learning (ASTL) for Offline BCI Calibration}

\author{Dongrui Wu\\
DataNova, NY USA\\
E-mail: drwu09@gmail.com}

\maketitle

\begin{abstract}
Single-trial classification of event-related potentials in electroencephalogram (EEG) signals is a very important paradigm of brain-computer interface (BCI). Because of individual differences, usually some subject-specific calibration data are required to tailor the classifier for each subject. Transfer learning has been extensively used to reduce such calibration data requirement, by making use of auxiliary data from similar/relevant subjects/tasks. However, all previous research assumes that all auxiliary data have been labeled. This paper considers a more general scenario, in which part of the auxiliary data could be unlabeled. We propose active semi-supervised transfer learning (ASTL) for offline BCI calibration, which integrates active learning, semi-supervised learning, and transfer learning. Using a visual evoked potential oddball task and three different EEG headsets, we demonstrate that ASTL can achieve consistently good performance across subjects and headsets, and it outperforms some state-of-the-art approaches in the literature.
\end{abstract}


\section{Introduction}

Single-trial classification of event-related potentials (ERPs) in electroencephalogram (EEG) signals has been used in many real-world brain-computer interface (BCI) systems \cite{Sajda2010,Bigdely-Shamlo2008,Zander2011}. However, because of large individual differences \cite{Bulayeva1993}, we need to calibrate the classifier for each subject, using some labeled subject-specific data. This paper investigates how to reduce the amount of labeled subject-specific calibration data.

As in our previous research \cite{drwuTHMS2017}, we distinguish between two types of calibration in BCI:
\begin{enumerate}
\item \emph{Offline calibration}: A pool of unlabeled EEG epochs from a subject have been obtained \emph{a priori}, and the subject is queried to label some of these epochs, which are then used to train a classifier to label the remaining epochs in that pool.
\item \emph{Online calibration}: Some labeled EEG epochs from a subject are obtained on-the-fly, and then a classifier is trained from them to classify future EEG epochs from the same subject.
\end{enumerate}
There are two main differences between these two calibration scenarios \cite{drwuTHMS2017}: 1) In offline calibration, the unlabeled EEG epochs can be used to help design the ERP classifier, whereas in online calibration there are no unlabeled EEG epochs; and, 2) in offline calibration we can query any epoch in the pool for the label, but in online calibration usually the sequence of the epochs is pre-determined.

This paper proposes a new offline calibration approach, by integrating regularization \cite{Scholkopf2001,Lotte2010}, active learning (AL) \cite{drwuRSVP2016,drwuEBMAL2016}, transfer learning (TL) \cite{Jayaram2016,drwuTNSRE2016}, and semi-supervised learning (SSL) \cite{Li2008,Chapelle2006}. We use a visual evoked potential (VEP) oddball task and three different EEG headsets to demonstrate that the proposed approach can achieve consistently good performance across subjects and headsets, and it outperforms our latest approach in \cite{drwuTHMS2017}. The experiment and dataset used in this paper are identical to those in \cite{drwuTHMS2017}; however, the approaches are different:
\begin{enumerate}
\item \cite{drwuTHMS2017} assumed data from every auxiliary subject are labeled, whereas in this paper data from some auxiliary subjects are unlabeled.
\item \cite{drwuTHMS2017} did not include AL in offline calibration.
\item This paper introduces a more sophisticated way to fuse the base TL models from multiple auxiliary subjects.
\end{enumerate}

The remainder of the paper is organized as follows: Section~\ref{sect:ASTL} introduces the details of the offline calibration algorithm. Section~\ref{sect:experiments} describes the experiment setup that is used to evaluate the performances of different algorithms. Section~\ref{sect:offline} presents performance comparison of different offline calibration algorithms. Finally, Section~\ref{sect:conclusions} draws conclusions.

\section{Active Semi-supervised Transfer Learning (ASTL) for Offline BCI Calibration} \label{sect:ASTL}

This section introduces the iterative active semi-supervised transfer learning (ASTL) algorithm for offline BCI classifier calibration, which integrates AL \cite{Settles2009}, weighted adaptation regularization (wAR) \cite{drwuTHMS2017}, and spectral meta-learner (SML) \cite{Parisi2014}.

\subsection{Problem Definition}

A \emph{domain} \cite{Pan2010,Long2014} $\mathcal{D}$ in TL consists of a feature space $\mathcal{X}$ and a marginal probability distribution $P(\mathbf{x})$, i.e., $\mathcal{D}=\{\mathcal{X},P(\mathbf{x})\}$, where $\mathbf{x}\in \mathcal{X}$. Two domains $\mathcal{D}_s$ and $\mathcal{D}_t$ are different if $\mathcal{X}_s\neq \mathcal{X}_t$, and/or $P_s(\mathbf{x})\neq P_t(\mathbf{x})$.

A \emph{task} \cite{Pan2010,Long2014} $\mathcal{T}$ in TL consists of a label space $\mathcal{Y}$ and a conditional probability distribution $Q(y|\mathbf{x})$. Two tasks $\mathcal{T}_s$ and $\mathcal{T}_t$ are different if $\mathcal{Y}_s\neq \mathcal{Y}_t$, or $Q_s(y|\mathbf{x})\neq Q_t(y|\mathbf{x})$.

Given a \emph{source domain} $\mathcal{D}_s$ with $n$ labeled samples, $\{(\mathbf{x}_1,y_1),..., (\mathbf{x}_n,y_n)\}$, and a \emph{target domain} $\mathcal{D}_t$ with $m_l$ labeled samples $\{(\mathbf{x}_{n+1},y_{n+1}),...,(\mathbf{x}_{n+m_l},y_{n+m_l})\}$ and $m_u$ unlabeled samples $\{\mathbf{x}_{n+m_l+1}, ..., \mathbf{x}_{n+m_l+m_u}\}$, \emph{domain adaptation} TL learns a target prediction function $f: \mathbf{x}_t \mapsto y_t$ with low expected error on $\mathcal{D}_t$, under the assumptions $\mathcal{X}_s=\mathcal{X}_t$, $\mathcal{Y}_s=\mathcal{Y}_t$, $P_s(\mathbf{x})\neq P_t(\mathbf{x})$, and $Q_s(y|\mathbf{x})\neq Q_t(y|\mathbf{x})$.

For example, in single-trial classification of VEPs studied in this paper, the source domain consists of EEG epochs from an existing subject, and the target domain consists of EEG epochs from a new subject. When there are $Z$ source domains, we perform domain adaptation TL for each of them separately and then aggregate the $Z$ classifiers. Because of individual differences, the source domain samples are usually not completely consistent with the target domain samples, and must be integrated with some labeled target domain samples to train an accurate target domain classifier.

Let $m_1$ and $m_2$ be the true number of target domain samples from Classes 1 and 2, respectively. Let $\hat{m}_1$ and $\hat{m}_2$ be the number of target domain samples that are correctly classified by ASTL as Classes 1 and 2, respectively. ASTL optimizes the following balanced classification accuracy (BCA):
\begin{align}
BCA=\frac{a_1+a_2}{2}\equiv \pi \label{eq:a}
\end{align}
where
\begin{align*}
a_1=\frac{\hat{m}_1}{m_1},\qquad a_2=\frac{\hat{m}_2}{m_2}
\end{align*}
i.e., $a_c$ is the classification accuracy for Class $c$, $c=1,2$.

\begin{figure}[htpb]\centering
\includegraphics[width=.9\linewidth]{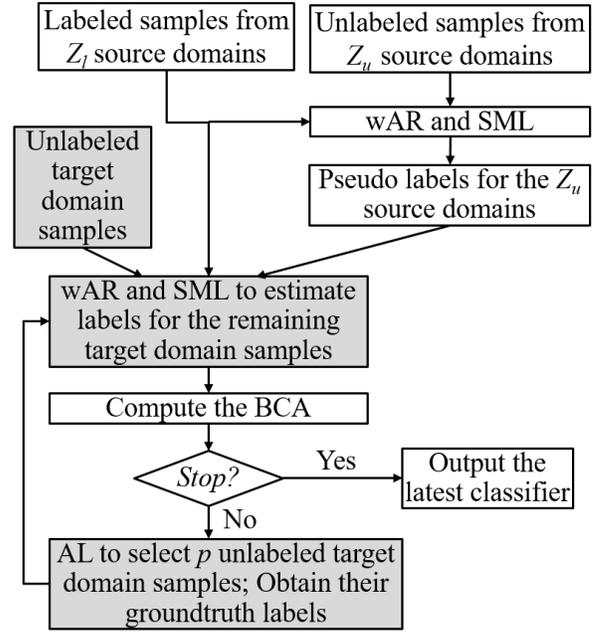}
\caption{ASTL for offline calibration.} \label{fig:ASTL}
\end{figure}

Assume there are $Z_l$ source domains with labeled samples, and $Z_u$ source domains with unlabeled samples. The flowchart of the iterative ASTL is shown in Fig.~\ref{fig:ASTL}. Based on the $Z_l$ labeled source domains, it first uses wAR and SML to estimate the pseudo labels separately for each of the $Z_u$ unlabeled source domains. Then, with the help of a few labeled target domain samples (selected by AL), ASTL uses wAR and SML on the $Z=Z_l+Z_u$ source domains to estimate the labels for the remaining target domain samples. We emphasize that ASTL uses SSL because: 1) it makes use of unlabeled samples in the $Z_u$ source domains in wAR and SML, and, 2) it also makes use of the unlabeled target domain samples in wAR.

The three main components of ASTL (wAR, SML, and AL) are introduced in more details next.

\subsection{Weighted Adaptation Regularization (wAR)} \label{sect:wAR}

The wAR algorithm has been introduced in our previous research \cite{drwuTNSRE2016,drwuTHMS2017} for classification. Recently it has also been extended to regression problems \cite{drwuTFS2017}.

Let the binary classifier be $f(\mathbf{x})=\mathbf{w}^T\phi(\mathbf{x})$, where $\mathbf{w}$ is the classifier parameters, and $\phi:\mathcal{X}\mapsto \mathcal{H}$ is the feature mapping function that projects the original feature vector to a Hilbert space $\mathcal{H}$. The learning framework of wAR is formulated as:
\begin{align}
f=&\argmin\limits_{f\in\mathcal{H}_K}\sum_{i=1}^{n}w_{s,i}\ell(f(\mathbf{x}_i),y_i)...
+w_t\sum_{i=n+1}^{n+m_l}\cdot w_{t,i}\ell(f(\mathbf{x}_i),y_i)\nonumber \\
&+\sigma\|f\|_K^2+\lambda [D_{f,K}(P_s,P_t)+ D_{f,K}(Q_s,Q_t)] \label{eq:f}
\end{align}
where $\ell$ is the loss function, $K\in \mathbb{R}^{(n+m_l+m_u)\times(n+m_l+m_u)}$ is the kernel matrix with $K(\mathbf{x}_i,\mathbf{x}_j)=\langle\phi(\mathbf{x}_i),\phi(\mathbf{x}_j)\rangle$, and $\sigma$ and $\lambda$ are non-negative regularization parameters. $w_t$ is the overall weight for target domain samples, which should be larger than 1 so that more emphasis is given to target domain samples than source domain samples. $w_{s,i}$ and $w_{t,i}$ are the weight for the $i^\mathrm{th}$ sample in the source domain and target domain, respectively, i.e.,
\begin{align}
w_{s,i}&=\left\{\begin{array}{ll}
                  1, & \mathbf{x}_i\in \mathcal{D}_{s,1} \\
                  n_1/(n-n_1), & \mathbf{x}_i\in \mathcal{D}_{s,2}
                \end{array}\right. \label{eq:ws}\\
w_{t,i}&=\left\{\begin{array}{ll}
                  1, & \mathbf{x}_i\in \mathcal{D}_{t,1} \\
                  m_1/(m_l-m_1), & \mathbf{x}_i\in \mathcal{D}_{t,2}
                \end{array}\right.  \label{eq:wt}
\end{align}
in which $\mathcal{D}_{s,c}=\{\mathbf{x}_i|\mathbf{x}_i\in \mathcal{D}_s\wedge y_i=c,\ i=1,...,n\}$ is the set of samples in Class $c$ of the source domain, $\mathcal{D}_{t,c}=\{\mathbf{x}_j|\mathbf{x}_j\in \mathcal{D}_t\wedge y_j=c,\ j=n+1,...,n+m_l\}$ is the set of samples in Class $c$ of the target domain, $n_c$ is the number of elements in $\mathcal{D}_{s,c}$, and $m_c$ is the number of elements in $\mathcal{D}_{t,c}$.

Briefly speaking, the 1st term in (\ref{eq:f}) minimizes the loss on fitting the labeled samples in the source domain, the 2nd term minimizes the loss on fitting the labeled samples in the target domain, the 3rd term minimizes the structural risk of the classifier, and the 4th term minimizes the distance between the marginal probability distributions $P_s(\mathbf{x}_s)$ and $P_t(\mathbf{x}_t)$, and also the distance between the conditional probability distributions $Q_s(\mathbf{x}_s|y_s)$ and $Q_t(\mathbf{x}_t|y_t)$.

Let
\begin{align}
X&=[\mathbf{x}_1, ...,\mathbf{x}_{n+m_l+m_u}]^T \label{eq:X} \\
\mathbf{y}&=[y_1,...,y_{n+m_l+m_u}]^T \label{eq:y}
\end{align}
where $\{y_1,...,y_n\}$ are known labels in the source domain, $\{y_{n+1},...,y_{n+m_l}\}$ are known labels in the target domain, and $\{y_{n+m_l+1},...,y_{n+m_l+m_u}\}$ are pseudo labels for the unlabeled target domain samples, i.e., labels estimated using available sample information in both source and target domains.

Define $E,M_0,M\in \mathbb{R}^{(n+m_l+m_u)\times(n+m_l+m_u)}$ as:
\begin{align}
E_{ii}&=\left\{\begin{array}{ll}
                w_{s,i}, & 1\le i\le n \\
                w_t\cdot w_{t,i}, &  n+1 \le i \le n+m_l\\
                0, & \text{otherwise}
              \end{array}\right. \label{eq:E}\\
(M_0)_{ij}&=\left\{\begin{array}{ll}
                             \frac{1}{n^2},& 1\le i \le n, 1\le j \le n  \\
                             \frac{1}{(m_l+m_u)^2}, & n+1\le i \le n+m_l+m_u,\\
                             & n+1\le j \le n+m_l+m_u \\
                             \frac{-1}{n(m_l+m_u)}, & \text{otherwise}
                           \end{array}\right. \label{eq:M0} \\
M&=M_1+M_2 \label{eq:M}
\end{align}
in which
\begin{align}
(M_c)_{ij}=\left\{\begin{array}{ll}
                    1/n_c^2, & \mathbf{x}_i, \mathbf{x}_j\in \mathcal{D}_{s,c} \\
                    1/m_c^2, & \mathbf{x}_i, \mathbf{x}_j\in \mathcal{D}_{t,c} \\
                    -1/(n_cm_c), & \mathbf{x}_i\in\mathcal{D}_{s,c}, \mathbf{x}_j\in \mathcal{D}_{t,c}, \\
                    &\text{or } \mathbf{x}_j\in\mathcal{D}_{s,c}, \mathbf{x}_i\in \mathcal{D}_{t,c} \\
                    0, & \text{otherwise}
                  \end{array}\right.
\end{align}
Then, according to \cite{drwuTNSRE2016,drwuTHMS2017}, the solution of $f(\mathbf{x})$ in (\ref{eq:f}) is:
\begin{align}
f(\mathbf{x})=\sum_{i=1}^{n+m_l+m_u}\alpha_iK(\mathbf{x}_i,\mathbf{x})=\boldsymbol{\alpha}^TK(X,\mathbf{x}) \label{eq:f2}
\end{align}
where
\begin{align}
\boldsymbol{\alpha}=[(E+\lambda M_0+ \lambda M)K+\sigma I]^{-1}E\mathbf{y} \label{eq:alpha}
\end{align}
The complete wAR algorithm is given in Algorithm~\ref{alg:wAR}.

\begin{algorithm}[h] 
\KwIn{$Z$ source domains, where the $z^\mathrm{th}$ ($z=1,...,Z$)\\
\hspace*{3.5em} domain has $n_z$ labeled samples\\
\hspace*{3.5em} $\{\mathbf{x}_i^z,y_i^z\}_{i=1,...,n_z}$\;
\hspace*{2.8em} $m_l$ labeled target domain samples,\\
\hspace*{3.5em} $\{\mathbf{x}_j^t,y_j^t\}_{j=1,...,m_l}$\;
\hspace*{2.8em} $m_u$ unlabeled target domain samples,\\
\hspace*{3.5em} $\{\mathbf{x}_j^t\}_{j=m_l+1,...,m_l+m_u}$\;
\hspace*{2.8em} Parameters $w_t$, $\sigma$ and $\lambda$ in (\ref{eq:f});}
\KwOut{The wAR classifier.}
Choose a kernel function $K(\mathbf{x}_i,\mathbf{x}_j)$\;
\For{$z=1,2,...,Z$}{
Construct the feature matrix $X$ in (\ref{eq:X})\;
Compute the kernel matrix $K$ from $X$\;
Set $y_j=\bar{y}_{j-n_z}^t$ ($j=n_z+m_l+1,...,n_z+m_l+m_u$), if $\bar{y}_{j-n_z}^t$ are available; otherwise, build another classifier (e.g., an SVM) to estimate them\;
Construct $\mathbf{y}$ in (\ref{eq:y}), $E$ in (\ref{eq:E}), $M_0$ in (\ref{eq:M0}), and $M$ in (\ref{eq:M})\;
Compute $\boldsymbol{\alpha}$ by (\ref{eq:alpha}) and record it as $\boldsymbol{\alpha}_z$\;
Use $\boldsymbol{\alpha}$ to classify the $n_z+m_l$ labeled samples and record the accuracy, $a_z$;}
\textbf{Return} $f(\mathbf{x})=\sum_{z=1}^Za_z\boldsymbol{\alpha}_zK_z(X,\mathbf{x})$.
\caption{The offline wAR algorithm \cite{drwuTNSRE2016,drwuTHMS2017}.} \label{alg:wAR}
\end{algorithm}

\subsection{Spectral Meta-Learner (SML)} \label{sect:SML}

SML is a simple yet effective way to optimally combine multiple base machine learning models. It was first proposed for binary classification problems \cite{Parisi2014}, and recently extended to regression problems \cite{drwuSMLR2016}.

Let $Q\in\mathbb{R}^{Z\times Z}$ be the population covariance matrix of $Z$ binary classifiers, i.e.,
\begin{align}
q_{zj}=\mathbb{E}\left[(f_z(\mathbf{x})-\mu_z)(f_j(\mathbf{x})-\mu_j)\right]
\end{align}
where $\mathbb{E}$ is the expectation, and $\mu_z=\mathbb{E}[f_z(\mathbf{x})]$. Parisi et al. \cite{Parisi2014} showed that:
\begin{align}
q_{zj}=\left\{\begin{array}{ll}
                1-\mu_z^2, & z=j \\
                (1-b^2)(2\pi_z-1)(2\pi_j-1), & z\neq j
              \end{array}\right. \label{eq:qij}
\end{align}
where $\pi_z$ is computed by (\ref{eq:a}) for each base classifier, and $b\in(-1,1)$ is the class imbalance, i.e.,
\begin{align}
b=\mathrm{Prob}(y\mbox{ in Class 1})-\mathrm{Prob}(y\mbox{ in Class 2})
\end{align}
(\ref{eq:qij}) shows that the off-diagonal entries of $Q$ are identical to those of a rank-one matrix $R=\lambda \mathbf{vv}^T$, where
\begin{align}
\left\{\begin{array}{rcl}
        \lambda&=&1-b^2\\
        \mathbf{v}&=& [2\pi_1-1, 2\pi_2-1,...,2\pi_Z-1]^T
       \end{array}\right.
\end{align}
In other words, the BCAs of the $M$ classifiers can be computed from the first leading eigenvector of $R$. Given the empirical $Q$, several different approaches \cite{Parisi2014} can be used to estimate $R$. The simplest is to use $Q$ directly as $R$, as in Algorithm~\ref{alg:SML}.

Once $\{\pi_z\}_{z=1,2,...,Z}$ are estimated, they can be used to replace $w_z$ in Algorithm~\ref{alg:wAR}, i.e.,
the following weighted average is used to aggregate the $M$ base binary classifiers:
\begin{align}
f(\mathbf{x})=\sum_{z=1}^Z\pi_z\boldsymbol{\alpha}_zK_z(X,\mathbf{x}) \label{eq:f2}
\end{align}

\begin{algorithm}[h] 
\KwIn{$m_u$ unlabeled samples, $\{\mathbf{x}_j\}_{j=1}^{m_u}$\;
\hspace*{2.8em} $Z$ binary classifiers, $\{f_z\}_{z=1}^Z$.}
\KwOut{The estimated BCAs of the $Z$ binary classifiers, $\{\pi_z\}_{z=1}^Z$.}
\For{$z=1,2,...,Z$}{
Compute $\{f_z(\mathbf{x}_j)\}_{j=1}^{m_u}$\;}
Compute the $Z\times Z$ covariance matrix $Q$ of $\{f_z(\mathbf{x})\}_{z=1}^Z$\;
Compute the first leading eigenvector, $\mathbf{v}$, of $Q$\;
\textbf{Return} $\pi_z=(v_z+1)/2$, $z=1,2,...,Z$, where $v_z$ is the $z$th element of $\mathbf{v}$.
\caption{The SML algorithm \cite{Parisi2014}.} \label{alg:SML}
\end{algorithm}

\subsection{Active Learning (AL)} \label{sect:AL}

Many AL approaches have been proposed in the literature \cite{Settles2009}. In ASTL we use the most popular idea:
we identify the $p$ most informative samples as the $p$ most uncertain ones, based on the latest classification boundary.

More specifically, we rank the uncertainties of the unlabeled target domain samples by their distance to the current classification boundary: a sample closer to the classification boundary means the classifier has more uncertainty about its class, and hence we should select it for labeling in the next iteration. To do this, we first sort the unlabeled target domain samples in ascending order according to $|f(\mathbf{x})|$ in (\ref{eq:f2}). Then, we select the first $p$ samples for labeling in the next iteration, as illustrated in Algorithm~\ref{alg:AL}.

\begin{algorithm}[h] 
\KwIn{$\{f(\mathbf{x}_j)\}_{j=1}^{m_u}$ in (\ref{eq:f2}).}
\KwOut{The $p$ most uncertain samples to label.}
Sort $\{|f(\mathbf{x}_j)|\}_{j=1}^{m_u}$ in ascending order\;
\textbf{Return} The first $p$ samples corresponding to the sorted $\{|f(\mathbf{x}_j)|\}_{j=1}^{m_u}$.
\caption{The AL algorithm.} \label{alg:AL}
\end{algorithm}

\section{The VEP Oddball Experiment} \label{sect:experiments}

This section describes the setup of the VEP oddball experiment, which is identical to that used in \cite{drwuTHMS2017}. It is used in the next three sections to evaluate the performances of different algorithms.

\subsection{Experiment Setup}

A two-stimulus VEP oddball task \cite{Ries2014} was used. Participants were seated in a sound-attenuated recording chamber, and image stimuli were presented to them at a rate of 0.5 Hz. The images (152$\times$375 pixels), presented for 150 ms at the center of a 24 inch Dell P2410 monitor at a distance of approximately 70 cm, were either an enemy combatant (target) or a U.S. Soldier (non-target). The subjects were instructed to maintain fixation on the center of the screen and identify each image as being target or non-target with a unique button press as quickly and accurately as possible. A total of 270 images were presented to each subject, among which 34 were targets. The experiments were approved by U.S. Army Research Laboratory Institutional Review Board. The voluntary, fully informed consent of the persons used in this research was obtained as required by federal and Army regulations \cite{USArmy,USDoD}. The investigator has adhered to Army policies for the protection of human subjects.

Signals for each subject were recorded with three different EEG headsets, including a 64-channel 512Hz BioSemi ActiveTwo system, a 9-channel 256Hz Advanced Brain Monitoring (ABM) X10 system, and a 14-channel 128Hz Emotiv EPOC headset. Complete data from 14 subjects were recorded and analyzed in this paper.

\subsection{Preprocessing and Feature Extraction}

The preprocessing and feature extraction method for all three headsets was the same, except that for ABM and Emotiv headsets we used all the channels, but for the BioSemi headset we only used 21 channels (Cz, Fz, P1, P3, P5, P7, P9, PO7, PO3, O1, Oz, POz, Pz, P2, P4, P6, P8, P10, PO8, PO4, O2) mainly in the parietal and occipital areas, as in \cite{drwuTHMS2017}.

EEGLAB \cite{Delorme2004} was used for EEG signal preprocessing and feature extraction. For each headset, we first band-passed the EEG signals to [1, 50] Hz, then downsampled them to 64 Hz, performed average reference, and next epoched them to the $[0, 0.7]$ second interval timelocked to stimulus onset. We removed mean baseline from each channel in each epoch and removed epochs with incorrect button press responses\footnote{Button press responses were not recorded for most subjects using the ABM headset, so we used all 270 epochs for them.}. The final numbers of epochs from the 14 subjects are shown in Table~\ref{tab:epoch}. Observe that there is significant class imbalance for every subject.

\begin{table*}[htpb] \centering \setlength{\tabcolsep}{1mm}
\caption{Number of epoches for each subject after preprocessing. The numbers of target epochs are given in the parentheses.}   \label{tab:epoch}
\begin{tabular}{l|cccccccccccccc}   \hline
   Subject  &  1&2&3&4&5&6&7&8&9&10&11&12&13 &14 \\ \hline
   BioSemi &  241 (26)&260 (24)& 257 (24) & 261 (29)& 259 (29)& 264 (30)& 261 (29) & 252 (22)& 261 (26)& 259 (29)& 267 (32)& 259 (24)&261 (25)& 269 (33)\\
  Emotiv  &263 (28) &  265 (30) &  266 (30)& 255 (23)& 264 (30)& 263 (32)& 266 (30)&252 (22)& 261 (26)& 266 (29)& 266 (32)& 264 (33) & 261 (26)& 267 (31)\\
  ABM & 270 (34) & 270 (34) & 235 (30) & 270 (34) & 270 (34)&270 (34)&270 (34)&270 (33)&270 (34)&239 (30)&270 (34)&270 (34)&251 (31)&270 (34)\\   \hline
\end{tabular}
\end{table*}

Each [0, 0.7] second epoch contains hundreds of raw EEG magnitude samples (e.g., $64\times0.7\times21=941$ for BioSemi). To reduce the dimensionality, we performed a simple principal component analysis (PCA) to take the scores on the first 20 principal components as features. We then normalized each feature dimension separately to $[0, 1]$.

\section{Performance Evaluation of Offline Calibration Algorithms} \label{sect:offline}

This selection presents performance comparison of ASTL with three other offline calibration algorithms.

\subsection{Offline Calibration Scenario}

Although we knew the labels of all EEG epochs for all 14 subjects in the VEP experiment, we simulated a realistic offline calibration scenario: we had labeled EEG epochs from 7 random subjects, unlabeled EEG epochs from another 6 random subjects, and also all epochs from the 14th subject, but initially none of those was labeled. Our goal was to iteratively label epochs from the 14th subject and build a classifier so that his/her remaining unlabeled epochs can be reliably classified.

Assume the 14th subject has $m$ unlabeled epochs, and we want to add $p$ ($p=5$ was used in this paper) labeled epochs in each iteration, starting from zero. In the first iteration, ASTL (see Fig.~\ref{fig:ASTL}) uses the $m$ unlabeled epochs from the 14th subject and epochs from the other 13 subjects to build an SML classifier and computes its BCA. It also uses AL to select the $p$ epochs that should be labeled next, and queries their labels. In the second iteration, ASTL builds a new classifier using the $p$ labeled epochs and $m-p$ unlabeled epochs from the 14th subject and epochs from the other 13 subjects, computes its BCA, selects another $p$ unlabeled target domain samples, and queries their labels. ASTL iterates until the maximum number of iterations is reached.

To obtain statistically meaningful results, the above process was repeated 30 times for each subject: each time we randomly selected 7 subjects to have labeled EEG epochs, and the remaining 6 subjects to have unlabeled EEG epochs. We repeated this procedure 14 times so that each subject had a chance to be the ``14th" subject.

\subsection{Offline Calibration Algorithms} \label{sect:Aoffline}

We compared the performance of ASTL in Fig.~\ref{fig:ASTL} with three other offline calibration algorithms \cite{drwuTHMS2017}:
\begin{enumerate}
\item \emph{Baseline 1 (BL1)}, which combines data from all 7 subjects with labeled epochs, builds a weighted support vector machine (SVM) classifier, and applies it to the target domain epochs. The SVM classifier does not update itself as new labeled target domain samples come in.
\item \emph{Baseline 2 (BL2)}, which is an iterative approach on available labeled target domain samples: in each iteration we randomly select 5 unlabeled samples from the 14th subject to label, and then train a weighted SVM classifier by 5-fold cross-validation. BL2 iterates until the maximum number of iterations is reached.
\item \emph{wAR} in Algorithm~\ref{alg:wAR}, on the 7 subjects with labeled epochs.  That is, it completely ignores the information in the unlabeled epochs from the other 6 subjects.
\end{enumerate}
Weighted libSVM \cite{LIBSVM} with radial basis function (RBF) kernel was used as the classifier in BL1 and BL2. The optimal RBF parameter was found by cross-validation. We chose $w_t=2$, $\sigma=0.1$, and $\lambda=10$, following the practice in \cite{drwuTHMS2017}.

\subsection{Offline Calibration Results} \label{sect:results}

The BCAs of the four algorithms, averaged over the 30 runs and across the 14 subjects, are shown in Fig.~\ref{fig:offline} for the three headsets. Observe that:
%

\begin{figure}[htpb]\centering
\subfigure[]{\label{fig:Boffline_avgs}     \includegraphics[width=.48\linewidth,clip]{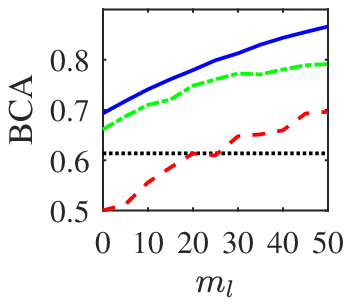}}
\subfigure[]{\label{fig:Aoffline_avgs}     \includegraphics[width=.48\linewidth,clip]{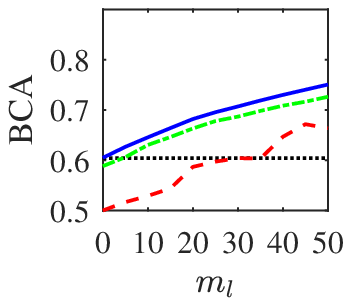}}
\subfigure[]{\label{fig:Eoffline_avgs}     \includegraphics[width=.72\linewidth,clip]{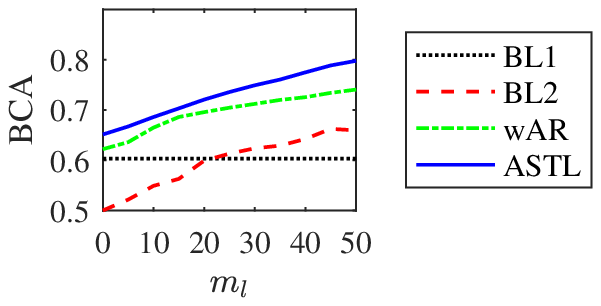}}
\caption{Average BCAs of the four \emph{offline} calibration algorithms across the 14 subjects, using different EEG headsets. (a) BioSemi; (b) ABM; (c) Emotiv. $m_l$ is the number of labeled samples from the ``14th" subject.} \label{fig:offline}
\end{figure}

\begin{enumerate}
\item Generally the performances of BL2, wAR and ASTL increased as more labeled subject-specific samples were available, which is intuitive.

\item BL2 cannot build a classifier when there were no labeled subject-specific samples at all (observe that the BCA for $m_l=0$ on the BL2 curve in Fig.~\ref{fig:offline} was always $0.5$, representing random guess), but wAR and ASTL can, because they can make use of information from other subjects. Moreover, without any labeled subject-specific samples, ASTL can build a classifier with a BCA of $69.31\%$ for BioSemi, $60.48\%$ for ABM, and $65.12\%$ for Emotiv, much better than random guess.

\item wAR significantly outperformed BL1 and BL2 (except for BL1 on ABM when $m_l$ is small), as demonstrated in our previous research \cite{drwuTHMS2017}.

\item ASTL significantly outperformed all the other three algorithms, except for BL1 on ABM when $m_l$ is small.
\end{enumerate}

As in \cite{drwuTHMS2017}, we also performed comprehensive statistical tests to check if the performance differences among BL2, wAR and ASTL (BL1 was not included because it is not iterative) were statistically significant. We used the area-under-performance-curve (AUPC) to assess overall performance differences among these algorithms. The AUPC is the area under the curve of the BCAs obtained at each of the 30 runs, and is normalized to $[0, 1]$. A larger AUPC value indicates a better overall classification performance.

We used Friedman's test \cite{Friedman1940}, a two-way non-parametric ANOVA where column effects are tested for significant differences after adjusting for possible row effects. We treated the algorithm type (BL2, wAR, and ASTL) as the column effects, with subjects as the row effects. Each combination of algorithm and subject had 30 values corresponding to 30 runs performed. Friedman's test showed statistically significant differences among the three algorithms for each headset ($df=2$, $p=0.00$).

Then, non-parametric multiple comparison tests using Dunn's procedure \cite{Dunn1961,Dunn1964} was used to determine if the difference between any pair of algorithms was statistically significant, with a $p$-value correction using the false discovery rate method \cite{Benjamini1995}. The results showed that the performances of ASTL were statistically significantly different from BL2 and wAR for each headset ($p=0.0000$ for all cases, except $p=0.0116$ for ASTL vs wAR on ABM).

Recall that the differences between ASTL and wAR are: i) ASTL makes use of information in the unlabeled samples from 6 additional subjects, whereas wAR does not; ii) ASTL uses SML to fuse the $Z$ classifiers, whereas wAR uses the training cross-validation accuracies as classifier weights in a weighted average; and, iii) ASTL uses AL to optimally select the unlabeled target domain samples to label, whereas wAR uses random selection. The significant performance improvement of ASTL over wAR demonstrated that our three modifications were very effective.

In summary, we have demonstrated that given the same number of labeled subject-specific training samples, ASTL can significantly improve the offline calibration performance. In other words, given a desired classification accuracy, ASTL can significantly reduce the number of labeled subject-specific training samples. For example, in Fig.~\ref{fig:Eoffline_avgs}, the average BCA of BL2 is $66.32\%$, given 50 labeled subject-specific training samples. However, to achieve that BCA, on average ASTL only need 5 samples, corresponding to $90\%$ saving of the labeling effort. Moreover, Fig.~\ref{fig:Eoffline_avgs} also shows that, without using any labeled subject-specific samples, ASTL can achieve similar performance as BL2 which uses 40 samples. Similar observations can also be made for the BioSemi and ABM headsets.

\section{Conclusions} \label{sect:conclusions}

Single-trial classification of ERPs in EEG signals has been used in many real-world BCI systems. Because of individual differences, usually some subject-specific calibration data are required to tailor the classifier for each subject. Many different approaches have been proposed for reducing such calibration data requirement, including regularization, TL, AL, SSL, etc, with TL being perhaps the most popular approach. TL makes use of auxiliary data from similar/relevant subjects/tasks. However, according to the author's knowledge, all previous applications of TL to BCI assume that all auxiliary data have been labeled. This paper considered a more general scenario, in which part of the auxiliary data could be unlabeled. We proposed ASTL for offline BCI calibration, which integrates regularization, AL, SSL, and TL. Using a VEP oddball task and three different EEG headsets, we demonstrated that ASTL can achieve consistently good performance across subjects and headsets, and it significantly outperformed wAR proposed in our recent research.

%
%
%


\end{document}